# Use of Student's *t*-Distribution for the Latent Layer in a Coupled Variational Autoencoder

Kevin R. Chen[1], Daniel Svoboda[2], and Kenric P. Nelson[3]

*Abstract: A Coupled Variational Autoencoder, which incorporates both a generalized loss function and latent layer distribution, shows improvement in the accuracy and robustness of generated replicas of MNIST numerals. The latent layer uses a Student's t-distribution to incorporate heavy-tail decay. The loss function uses a coupled logarithm, which increases the penalty on images with outlier likelihood. The generalized mean of the generated image's likelihood is used to measure the performance of the algorithm's decisiveness, accuracy, and robustness.*

## 1) Introduction: Variational Autoencoders

The *variational autoencoder* (VAE) (Kingma and Welling, 2014 & 2019) is a hybrid unsupervised model that combines probabilistic programming with deep learning. Its multi-layered autoencoder is augmented with the methods of Bayesian variational inference (Blei, et al., 2017), including a probabilistic latent layer and an information-theoretic loss function. Within the VAE, its recognition model or encoder $q_\Theta(z|x)$ first outputs a Gaussian probability density function, made up of mean ($\mu$) and standard deviation ($\sigma$) latent vectors. It then samples q to create the *probabilistic approximation* of the latent layer **z**. Finally, its generative model or decoder $p_\phi(x'|z)$ inputs **z** to generate the output sample x'.

The training optimizes the sum of two loss functions: *Kullback-Leibier Divergence,* which measures the distance between the latent distribution and a prior distribution (typically a standard normal), and *cross-entropy*, which measures the data reconstruction loss. By regularizing the latent layer with a probability distribution, the VAE can more tightly control its encoder $\Theta$ and decoder $\phi$ parameters during the learning process.

As increasingly sophisticated deep learning approaches continue to integrate into complex "real-world" applications, from robotic autonomous systems to physical and biomedical simulations, the features generated by the processes are often non-linear, leading to heavy-tailed distributions. Unfortunately, these heavy-tails add to the mathematical complexity and noise. As existing probabilistic programming methods typically assume properties of exponential tail decay (i.e., Gaussian distributions, Shannon information theory, and mean and variance estimations), their effectiveness in regards to deep learning models, such as VAE, can be impacted in unexpected ways if the data generating process is not well approximated by a multivariate normal distribution.

---

[1] Lead Author: kevinrchen0@gmail.com
[2] Principal Software Developer: daniel.svoboda@gmail.com
[3] Corresponding Author: kenric.nelson@gmail.com, Photrek, LLC

## 2) Coupled VAE: Modeling and reducing the impact of outliers

*Nonlinear Statistical Coupling* (NSC) (Nelson and Umarov, 2010; Nelson et al., 2017) provides a suite of techniques to incorporate modeling of heavy-tail phenomena into probabilistic programming methods. The *coupled VAE,* introduced in (Cao et al., 2019), demonstrates the performance improvements in the *accuracy* and *robustness* which can be achieved. The three principal methods that NSC can address are the following:
1) Generalized logarithm for loss functions and performance metrics (Cao et al., 2019)
2) Generalized exponential for heavy tail latent layer distribution, as introduced in this report
3) Design of neural networks (Nelson et al., 2014), planned for future research

The first version of a coupled VAE implemented the generalized logarithm for loss functions and the generalized mean for performance metrics. Using the MNIST handwritten numeral dataset, this coupled VAE already shows that by increasing the coupling loss hyperparameter $\kappa_L$ up to 0.2, the standard deviation of the latent variable **z** shrinks. This indicates that $\kappa_L$ can act as a regularizer that imposes a severe penalty for the presence of outliers in either the latent distribution or the generated output image. This leads to higher likelihoods of approximating the actual input image **x** and therefore sharper, less-fuzzy images. Histogram analysis of the test-set likelihoods showed improvement in both the geometric mean (accuracy or central tendency) and the -⅔ mean (robustness or outlier suppression) metrics.

In this presentation, we report on the second version of coupled VAE, which includes the generalized logarithmic loss function from the first version while adding on the generalized latent layer distribution, using the Student's *t*-distribution. This further enhances the flexibility for designing the properties of the coupled VAE as it can adjust the heaviness of the tails on its latent layer distributions.

## 3) Student's t (coupled Gaussian) latent layer distribution

The Student's *t*-distribution, first conceived by the Guiness Brewery statistician William Gosset (Gosset, 1904), estimates the mean $\mu$ of presumably normal population distribution when a) the sample size is small and b) the population standard-deviation $\sigma$ is relatively unknown. Like the normal distribution, the Student's *t* is bell-shaped and symmetric. However, the key difference is the degree of freedom (DoF), which is equal to one less than the number of observations, controls the shape or "heaviness" of the tail decay. The smaller the DoF, the heavier the tail shape, and vice versa until the Student's *t* converges towards a normal distribution as DoF approaches infinity. With the DoF as a hyperparameter (sometimes denoted as **v**), the Student *t*-distribution provides a more accurate and robust model than the Gaussian.

The requirements to manage risk and uncertainty in complex systems, broadens the relevance of Student's *t*-distribution to situations in which either fluctuation in the variance or a risk-biased preference for robustness requires modeling of slow decaying tails. To unify these requirements with the role of NSC for generalizing the loss functions and performance metrics, we parameterize the Student's *t* with the inverse of the DoF, or coupling distribution $\kappa_D = 1/\text{DoF}$. The coupling parameter quantifies both the relative variance of the variance (fluctuations) and in turn the shape of the tail decay. For this reason, the Student's *t* is also referred to as the *coupled Gaussian*.

In the VAE, **z** is derived from a reparameterization function $g(\phi, x, \varepsilon)$ (Kingma and Welling, 2019), where the noisy random variable **ε** is sampled from a standard normal distribution p(ε). In this version's coupled VAE, we modify p to be a Student's *t* or coupled Gaussian distribution instead. Increasing the coupling distribution $\kappa_D$ leads to heavier tails of *g*, therefore broadening the variability of latent layer **z**. Conversely, as $\kappa_D \to 0$, the tail converges to the exponential decay of a Gaussian distribution. As we show in the next section, the $\kappa_D$ hyperparameter improves the modeling of uncertainty in the target dataset.

## 4) Experimental Method

In order to understand how the latent layer tail shape or coupling $\kappa_D$ affect the performance of the VAE, we build upon the experimental setup of the previous version of the coupled VAE, which only reported on the role of the coupling loss hyperparameter $\kappa_L$. That includes using the same MNIST handwritten digit database, which has a training set of 60,000 images and a test set of 10,000 images. Each image has a 28x28 dimension with a grey-scale coloring: pixel values ranging from 0 (white) to 255 (black). We also maintain the following model architecture and hyperparameters:
- Encoder: 2 dense networks
- Decoder: 2 dense networks
- Number neural units: 500 (all layers in encoder and decoder)
- Dropout rate: 0.1 (both encoder and decoder)
- Learning rate: $1e^{-3}$
- Batch size: 5,000
- Number epochs: 500

After discussing the similarities between the first and current versions of the coupled VAE, here are the following modifications we made between the two versions. First, as discussed in the Student's *t* section, we modify p(ε) from normal to Student's *t* distribution, with the coupling distribution hyperparameter $\kappa_D$ to inversely control the DoF or the heaviness of the tails. We use the `MultivariateStudentTLinearOperator` class from the `tensorflow_probability.distributions` library to implement the Student's *t*. Second, we present the coupled VAE results with the dimensions of the **z** latent layer (**z_dim**) set to both 2D (summarized in the next section) and 20D (planned for presentation). Although the 2D latent layer has a lower performance than 20D, it simplifies the illustration of the latent layer manifold.

Using the seed of 0, we test and plot the following combinations of $\kappa_D$ and $\kappa_L$ values to compare the performance of *robustness*, *accuracy*, and *decisiveness* metrics as well as to examine the generated image qualities. The three metrics are powers (*r*) of the generalized mean of the likelihood that a generated image matches the input test image, respectively:
- $r = -2/3$ for robustness,
- $r = 0$ for accuracy, which is the geometric mean, and
- $r = 1$ for decisiveness, which is the arithmetic mean

For the two coupling hyperparameters, 0 is the control value, specifying the original VAE design.

Setting both $\kappa_D = 0$ and $\kappa_L = 0$[4] is equivalent to the original VAE while setting only the $\kappa_D = 0$ is equivalent to the first version of the coupled VAE. For both $\kappa_D$ and $\kappa_L$, we have set the second value to 0.15 in order to give as much distribution between the two values while minimizing the risk of the coupled VAE not being able to converge.

## 5) Result and evaluation

For our results, we show the effect of the hyperparameters coupling distribution $\kappa_D$ and coupling loss $\kappa_L$ has on the shape of the latent space manifold (Figure 1), the likelihood histogram of the generated images (Figure 2 and Table 1), and the visual quality of the images (Figure 3). The histograms highlight the *robustness*, *accuracy*, and *decisiveness* metrics. The results indicate that improvement by multiple orders of magnitude is possible for the robustness metric.

The latent plots in Figure 1 show a much wider spread when the coupling distribution is 0.15. The clusters have both greater separation but also more dispersion. The histogram analysis in Figure 2 indicates that this improves the decisiveness, but limits the improvement in the accuracy. For instance, the dispersion of the number *5* is particularly severe for the $\kappa_D$: 0.15, $\kappa_L$: 0.15 case, and as Figure 3 shows the number *5* is not reproduced well. Another problematic case is distinguishing the proximity of *4* and *7* with *8* and *9*. Further improvement in the control of the latent layer dispersion is planned by a) modifying the prior probability of the Kullback-Leibler divergence from a standard normal distribution to a standard coupled normal distribution with a matching tail decay, and b) experimenting with negative coupling (or DoF) to tighten rather than broaden the dispersion within clusters, while still maintaining separation between clusters.

The original and baseline VAE has $\kappa_D = 0$, a Gaussian latent distribution, and $\kappa_L = 0$, a natural logarithm for the loss functions. Table 1 shows the following: the best decisiveness result is found at $\kappa_D$: 0.15, $\kappa_L$: 0, with a value of 7.16e-9. The best accuracy result is found at $\kappa_D$: 0, $\kappa_L$: 0.15 with a value of 5.43e-44. Finally, the best robustness result is found at $\kappa_D$: 0.15, $\kappa_L$: 0.15 with a value of 5.73e-128.

In conclusion, we show that hyperparameter tuning the combination of the coupling distribution and coupling loss (or either one) can improve all three of the generalized mean performance metrics: *decisiveness*, *accuracy*, and *robustness*. The current results are for a 2-dimensional latent layer, but our research plans include experimentation with higher dimensional latent spaces and evaluating a wider variety of $\kappa_D$ and $\kappa_L$ values in order to determine the best balance between higher performance metrics and maintaining stable model convergence.

---

[4] Implemented as $10^{-6}$ in our experiments to approximate the original loss function.

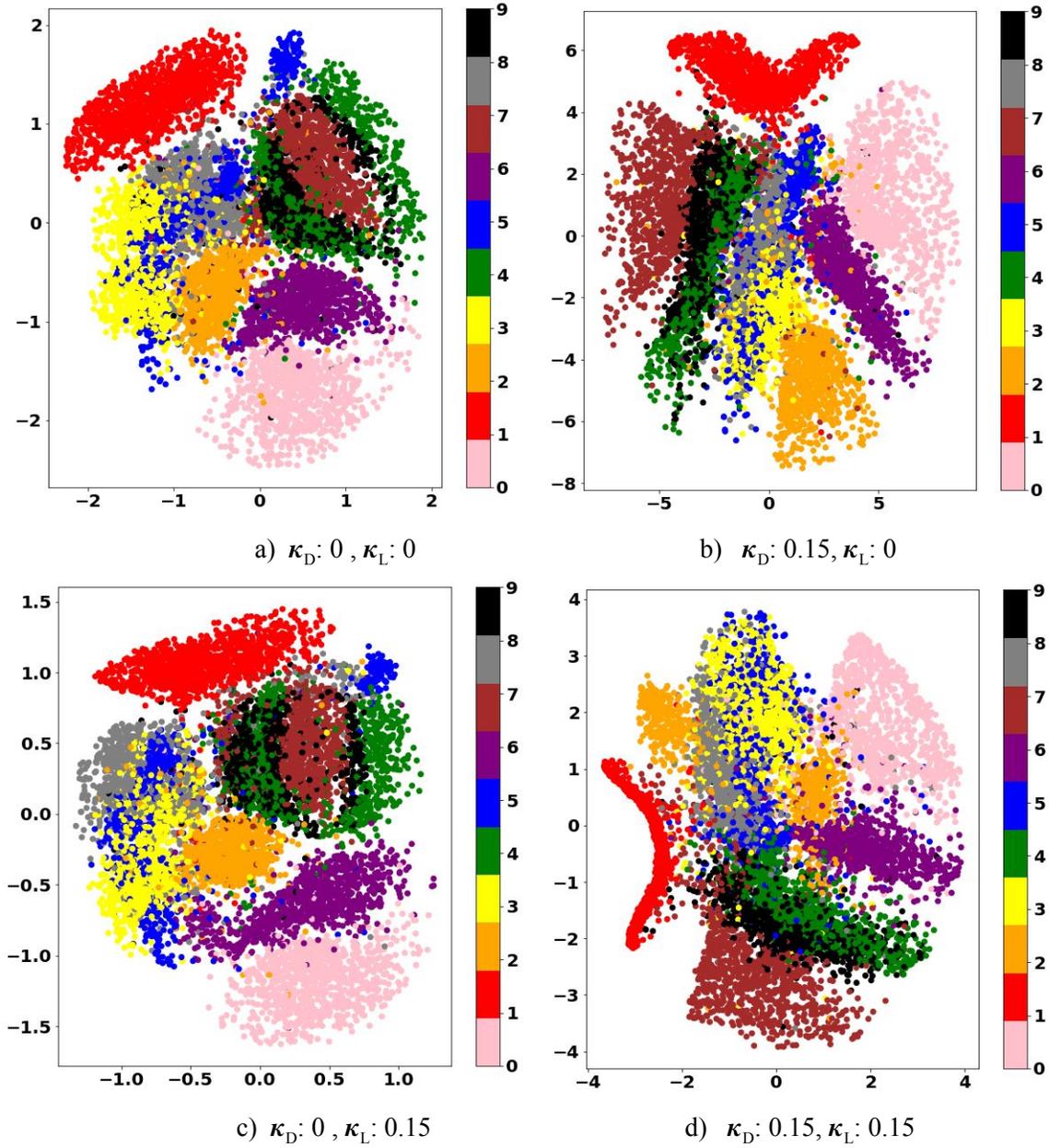

Figure 1: Plots of 2-D latent layer manifold for various cases of $\kappa_D$ and $\kappa_L$. These plots are generated by the latent layer as an output of the encoder input. Here, we take the mean vector that is generated in the latent layer and plot it on a 2-D graph. Each color represents a digit value 0 through 9. An item to notice is that for coupling distribution of 0.15, the clusters are spread out over a wider area.

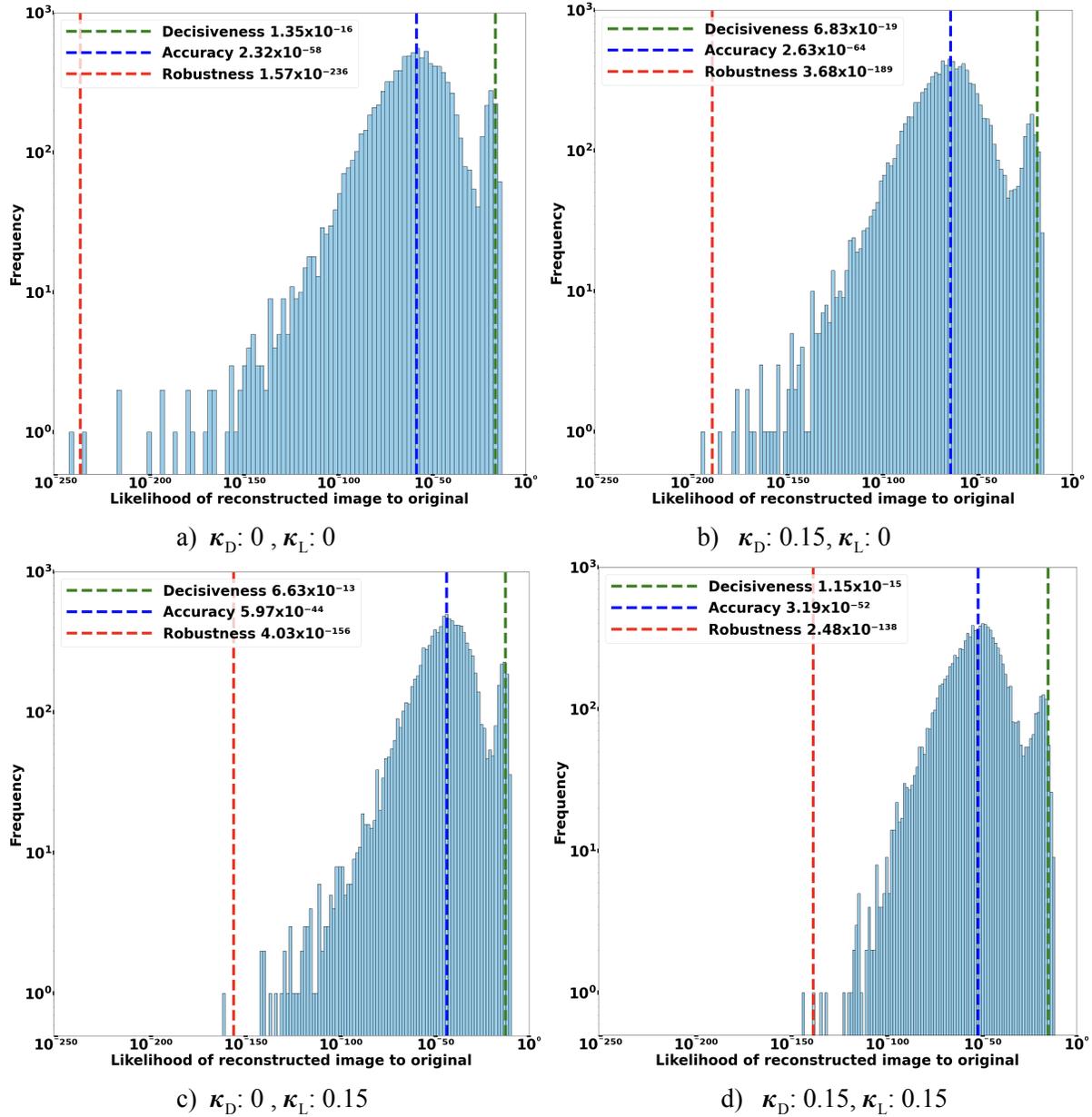

Figure 2: Histogram plots for various cases of $\kappa_D$ and $\kappa_L$. The histograms show the frequency of images with a given likelihood of the reconstruction. The red dotted line represents the computed robustness, blue is the accuracy and green is the decisiveness.

Table 1. Performance Metrics for Decisiveness, Accuracy, and Robustness

| $\kappa_D$ | $\kappa_L$ | *Decisiveness* | *Accuracy* | *Robustness* |
|---|---|---|---|---|
| 0 | 0 | $1.27e^{-16}$ | $2.25e^{-58}$ | $1.24e^{-258}$ |
| 0.15 | 0 | **$7.16e^{-9}$** | $2.29e^{-64}$ | $2.04e^{-189}$ |
| 0 | 0.15 | $6.57e^{-13}$ | **$5.43e^{-44}$** | $2.03e^{-159}$ |
| 0.15 | 0.15 | $4.17e^{-16}$ | $3.67e^{-52}$ | **$5.73e^{-128}$** |

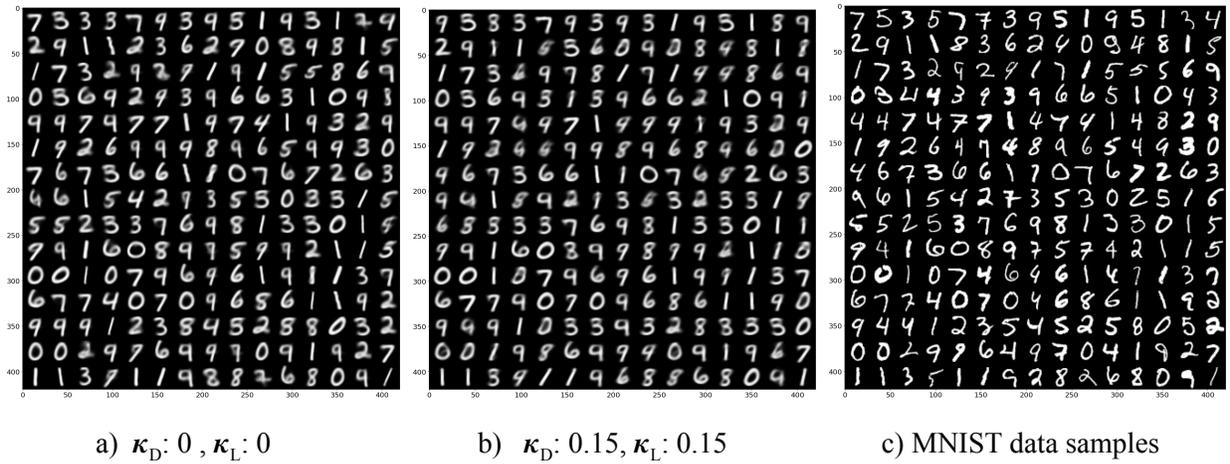

a) $\kappa_D$: 0, $\kappa_L$: 0    b) $\kappa_D$: 0.15, $\kappa_L$: 0.15    c) MNIST data samples

Figure 3: MNIST generated plots for various cases of $\kappa_D$ and $\kappa_L$. a) and b) represent generated images produced by the coupled VAE for respectively the original design and for when both coupling distribution and coupling loss are 0.15. Both images can be contrasted with c) samples from the original MNIST images that were inputted into the coupled VAE.